\documentclass{article}

\usepackage{spconf,amsmath,epsfig}
\usepackage{graphicx}
\usepackage{booktabs}
\usepackage{tikz}
\usepackage{comment}
\usepackage{amsfonts}
\usepackage{color}
\usepackage{hyperref}
\usepackage{bm}
\usepackage{wrapfig}
\usepackage{epsfig}
\usepackage{tabularx}
\usepackage{setspace}
\usepackage{caption}

\let\OLDthebibliography\thebibliography
\renewcommand\thebibliography[1]{
  \OLDthebibliography{#1}
  \setlength{\parskip}{0pt}
  \setlength{\itemsep}{0pt plus 0.3ex}
}

\begin{document}\sloppy

\def\x{{\mathbf x}}
\def\L{{\cal L}}

\title{Single Stage Warped Cloth Learning and Semantic-Contextual Attention Feature Fusion for Virtual TryOn}
%

\name{Sanhita Pathak$^{1}$, Vinay Kaushik$^{2}$, and Brejesh Lall$^{3}$}

\address{
\textit{$^{1}$Bharti School of Telecommunications, Indian Institute of Technology, Delhi, India} \\
\textit{$^{2}$Department of Computer Science, Indian Institute of Information Technology, Sonepat, Haryana, India} \\
\textit{$^{3}$Department of Electrical Engineering, Indian Institute of Technology, Delhi, India} \\
$^{1}$Sanhita.Pathak@dbst.iitd.ac.in, $^{2}$vkaushik@iiitsonepat.ac.in, $^{3}$brejesh@ee.iitd.ac.in
}


\maketitle

\begin{abstract}
Image-based virtual try-on aims to fit an in-shop garment onto a clothed person image. Garment warping, which aligns the target garment with the corresponding body parts in the person image, is a crucial step in achieving this goal. Existing methods often use multi-stage frameworks to handle clothes warping, person body synthesis and tryon generation separately or rely on noisy intermediate parser-based labels. We propose a novel single-stage framework that implicitly learns the same without explicit multi-stage learning. Our approach utilizes a novel semantic-contextual fusion attention module for garment-person feature fusion, enabling efficient and realistic cloth warping and body synthesis from target pose keypoints. By introducing a lightweight linear attention framework that attends to garment regions and fuses multiple sampled flow fields, we also address misalignment and artifacts present in previous methods. To achieve simultaneous learning of warped garment and try-on results, we introduce a Warped Cloth Learning Module. Our proposed approach significantly improves the quality and efficiency of virtual try-on methods, providing users with a more reliable and realistic virtual try-on experience.
\end{abstract}
\section{Introduction}
\label{sec:intro}

Virtual try-on technology has gained significant traction in the retail industry, offering a realistic and reliable experience for customers to virtually try on garments in-shop on their own images.
Among the different areas within virtual try-on, garment try-on has garnered substantial research interest being extensively explored in recent years \cite{minar2020cp,Issenhuth2020DoNM,fang2023pg,Choi2021VITONHDHV}. 

Due to the complexity of the virtual try-on problem, preserving both human and garment texture in varying body scenarios is crucial. Over the years, numerous methods have been proposed to address this challenge. VITON \cite{Han2017VITONAI}, one of the pioneering works in this field, utilized TPS warping to align garments with the human body and synthesized coarse results. CPVTON \cite{wang2018toward} and CPVTON+ \cite{minar2020cp} further improved the results by introducing semantic category refinement. OVITON \cite{Neuberger_2020_CVPR} proposed an online optimization module for appearance refinement. Building on these advances, \cite{ge2021disentangled} introduced disentanglement of garments and incorporated cycle consistency into self-supervised training.

Flow-based models, such as Flow-Style \cite{He2022StyleBasedGA} and DAFlow \cite{bai2022single}, used flow estimation to warp garments effectively.


Researchers have also leveraged prior knowledge, such as UV correspondence, to handle unobserved appearances. TryOnGAN \cite{adavala2023generation} utilized the StyleGAN2 generation method to improve body shape deformation. Further enhancing resolution , VITON-HD \cite{Choi2021VITONHDHV} and Dresscode \cite{fang2023pg} were introduced. WUTON \cite{Issenhuth2020DoNM} adopted a student-teacher model for appearance flow distillation, further pushing the boundaries of virtual try-on techniques.


Most try-on techniques follow a two stage tryon process (garment warp , tryon). In the context of garment warping, three primary techniques have been employed: Spatial Transformer Network (STN) \cite{NIPS2015_33ceb07b}, which learns affine transformations for rigid garment warping; Thin Plate Spline (TPS) deformation \cite{minar2020cp}, used for non-rigid garment warping by manipulating control points; and optical flow-based methods \cite{bai2022single}, which offer high-intensity deformable optical flow, estimating pixel offsets.

\begin{figure*}
    \centering
    \resizebox{\textwidth}{!}{\includegraphics{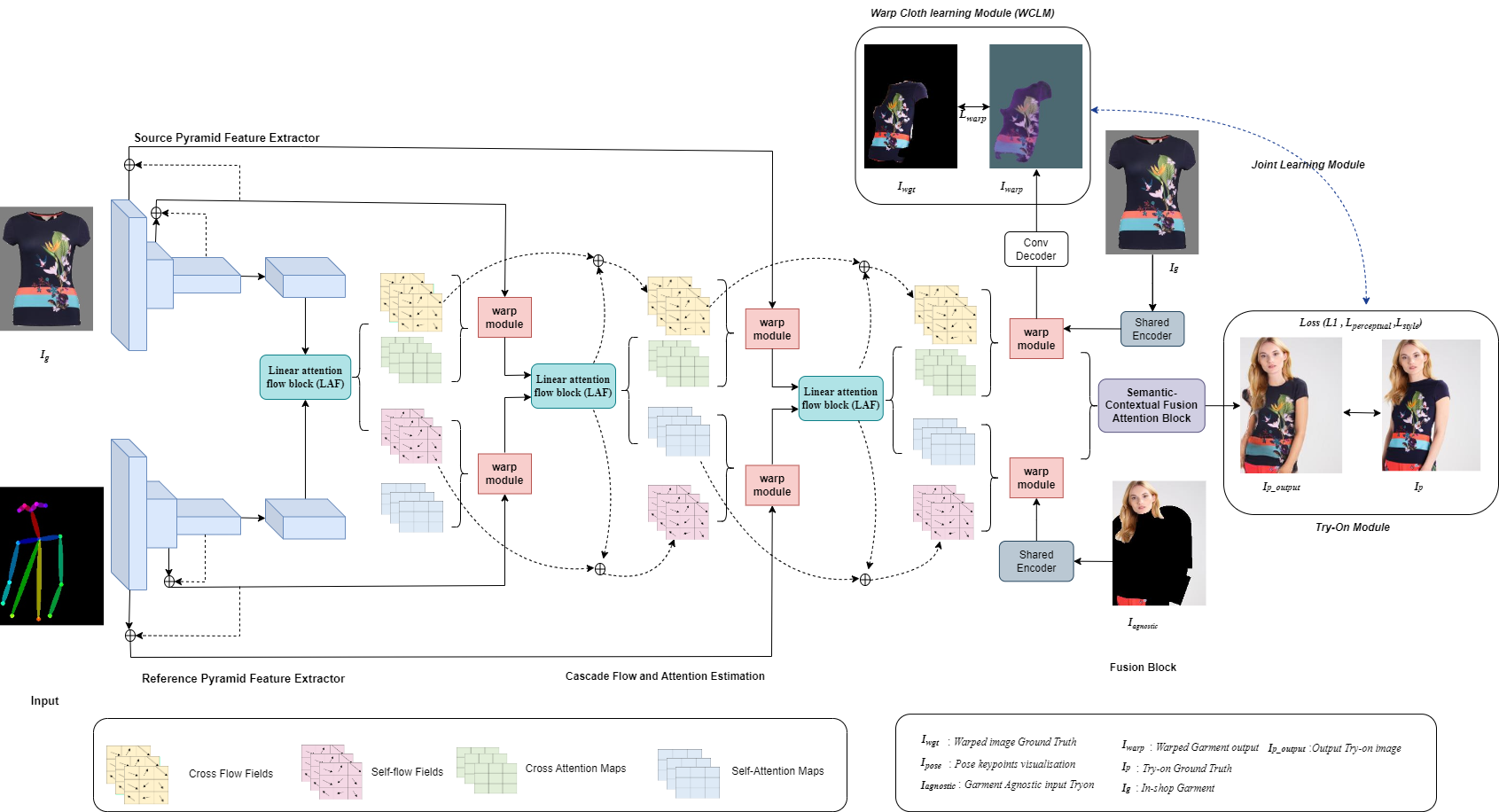}}
    \caption{The overall architecture framework of our method. The person, garment and pose is fed into the feature extractor followed by the LAF flow estimation. The joint learning module comprising of WCLM and tryon learning uses the resultant warped and tryon predictions for loss computation}
    \label{fig:q2}
\end{figure*}

Initially, many virtual try-on methods utilized the TPS warping method for input garment warping, leveraging its control points to align and non-rigidly warp the garment using an energy function. However, this method has been replaced by Flow-based warping techniques.
Ge et al. \cite{ge2021disentangled} disentangled the garment and non-garment regions, and Bai et al. \cite{bai2022single} proposed Flow-Style and DAFlow to warp garments using flow estimation. Furthermore, some studies combined pose transfer with virtual try-on, Dong et al. \cite{dong2019towards} proposed MGTON, which realized pose-guided virtual try-on. Xie et al. \cite{adavala2023generation} introduced PASTA-GAN to implement patch-routed disentanglement in unsupervised training. 
\par Recent advancements in self-attention mechanisms in vision transformers (ViTs) \cite{zamir2022restormer}, Swin Transformer \cite{zamir2022restormer},  Focal Transformer \cite{zamir2022restormer} and Restomer \cite{zamir2022restormer} have shown promise in various computer vision tasks which inspired the techniques to explore attention in tryon.
One such work that utilized attention in virtual try-on is PG-VTON \cite{fang2023pg}, which leverages attention to identify crucial regions in the image where the clothing makes contact with the body. Similarly, FashionGAN+ \cite{adavala2023generation} employs attention to transfer the appearance of clothing from one person to another while SDAFN (DAWarp) \cite{bai2022single}, focuses on warping a garment onto a subject's body while considering the subject's pose.

While single-stage frameworks for virtual try-on exist \cite{bai2022single}, they often lack an extra flow network for warping and instead apply an end-to-end try-on pipeline with deformable flow. Nevertheless, these methods may encounter issues with disproportionate warping, as the learning is not task-specific like in multiple-stage methods. Additionally, some approaches introduce heavy priors for the model in the form of parsing or densepose.
\par Our understanding of gaps is three fold. 
Most virtual tryon solutions involve an explicit garment warp based learning module which focuses on generating tryon image utilising a multistage pipeline. This incorporates sequential learning of intermediate representations such as optical flow or TPS transformation to compute warped garment as a separate deep learning module. The Optical flow calculation is done similar to corresponding rigid body paradigms, where inherent pixel offset calculation is done without using semantics or attention as guidance, giving equal importance to foreground and background regions. Also the blending multiple output features/images using a shallow decoder or formulate a complex  generative model as a separate learning framework causes misalignment and texture artifacts.

\par Our work makes significant contributions in the following aspects:
\begin{itemize}
\item A novel warped garment learning module that jointly learns the warped garment, person synthesis and final tryon result, explicitly as a single stage learning process.
\item A lightweight linear attention framework that attends to garment regions and fuses multiple sampled flow fields to learn optimal implicit garment flow.
\item We attain state-of-the-art results on the VITON dataset, demonstrating the effectiveness of our parser free single-stage virtual try-on approach.
\end{itemize}

\section{METHODOLOGY}

\subsection{Problem Definition}
The objective of virtual try-on is to produce a try-on image $I_{p\_output} \in \mathbb{R}^{3XHXW}$ by leveraging a person image $I_p \in \mathbb{R}^{3XHXW}$ and an in-shop garment image $I_g \in \mathbb{R}^{3XHXW}$. The aim is to ensure that the garment in $I_g$ seamlessly aligns with the corresponding regions in $I_p$, creating a coherent visual outcome. Moreover, the generated $I_{p\_{output}}$ should retain both the intricate elements from $I_g$ and the non-garment sections of $I_p$. In simpler terms, the individual portrayed in $I_p$ should remain unaltered in $I_{p\_{output}}$ except for the addition of the worn garment $I_g$.

Our proposed methodology encompasses several key components: An Attention-based Flow Estimator, Semantic-Contextual Fusion Attention Block, and a Joint Learning Tryon Module.

The Inputs to our method are the in-shop garment $I_g$ and cloth agnostic input $I_{agnostic}$ of subject person $I_p$ for tryon. Additionally, we utilize a keypoint visualization $I_{pose}$ to facilitate target person synthesis. 

Our method utilises two feature extractors. The source input is $I_g$ whereas the reference input is the channel concatenated $I_{agnostic}$ and $I_{pose}$. A Refined Pyramid Extractor operates by computing features at multiple scales \cite{bai2022single}. 

\begin{figure*}
    \centering
    \resizebox{\textwidth}{!}{\includegraphics{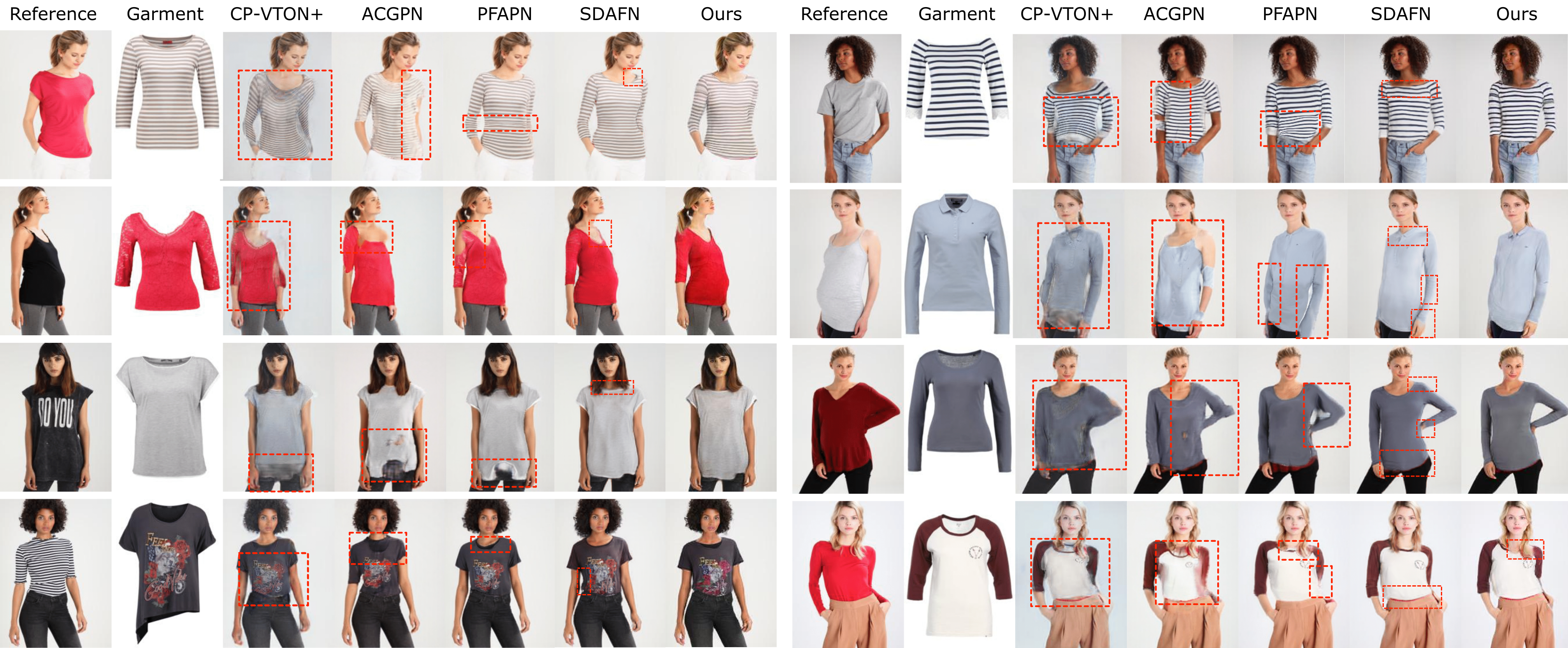}}
    \caption{Qualitative Results}
    \label{fig:q1}
\end{figure*}

These features are then employed for learning deformable flow through our innovative Linear Attention Flow (LAF) block. LAF learns both attention and flow across various scales, which is used by our warp module to compute warped features. These warped features are fed into the Semantic-Contextual Feature Fusion block to predict the final tryon image$I_{p\_{output}}$. It's noteworthy that attention and flow are learned using a multi-scale residual approach similar to Fig\ref{fig:q1}.

In the following subsections, we describe every module in detail.

\subsection{Linear Attention Flow Block (LAF)}

Transformers, pivotal in NLP and computer vision, rely on Query, Key, and Value (QKV) formulation. Despite enabling intricate dependency modeling, its quadratic complexity poses computational challenges. Addressing this, our proposed Linear Attention Flow Block (LAF) employs efficient linear attention, as in Wang et al. \cite{zamir2022restormer}. This alternative minimizes costly matrix multiplications, significantly reducing computation and memory requirements, ensuring LAF's scalability and efficiency at high spatial resolutions.

The fundamental principle of our proposed linear self-attention involves the incorporation of two linear projection matrices, $E_i$ and $F_i \in \mathbb{R}^{n\times k}$, during key and value computation. Initially, we project the original $(n\times d)$-dimensional key and value layers, $KW_i^K$ and $VW_i^V$, into $(k \times d)$-dimensional projected counterparts. Subsequently, the context mapping matrix $\bar{P}$ of dimensions $(n\times k)$ is derived through scaled dot-product attention, as illustrated by the equation:

\begin{align}
\overline{\text{head}_i} &= \text{Attention}(QW_i^Q, E_iKW_i^K, F_iVW_i^V)\notag\\
&=\underbrace{\text{softmax}\left(\frac{QW_i^Q(E_iKW_i^K)^T}{\sqrt{d_k}}\right)}_{\bar{P}: n\times k}\cdot\underbrace{F_iVW_i^V}_{k\times d},\label{eq:linearattenion}
\end{align}

The final step entails generating context embeddings for each head$_i$ utilizing $\bar{P}\cdot(F_iVW_i^V)$. Importantly, these operations exhibit a time and space complexity of $O(nk)$, which is considerably efficient. This efficiency becomes more pronounced when we opt for a small projected dimension $k$, substantially reducing memory and space requirements. Our LAF block stands as a testament to the balance struck between computational efficiency and robust attention modeling, making it well-suited for our Virtual Tryon pipeline. Finally, the attended values are obtained as $Output=Attention_{scores} \cdot V$.

\subsection{Semantic-Contextual Fusion Attention}
In this subsection, we propose a novel approach for computing the final try-on result in our virtual try-on pipeline. The baseline feeds pixel-wise summation of the features of the warped garment and the features of the generated reference person to a 2-layer convolutional shallow decoder to compute the target tryon result. This approach lacks the ability to effectively capture the fine-grained relationships between the two sets of features, potentially leading to suboptimal try-on results. To address this limitation, we introduce a Semantic-Contextual Fusion Attention (SCFA) module that encapsulates the attention mechanism's capability to capture semantic information while considering the context between the garment and the person during the try-on process. 

\subsubsection{Key-Value Attention Computation}
Given the features of the warped garment $F_{\text{garment}} \in \mathbb{R}^{C \times H \times W}$ and the features of the generated reference person $F_{\text{person}} \in \mathbb{R}^{C \times H \times W}$, we compute the KV attention as follows. We first apply a convolutional layer to create embeddings $E_{\text{garment}}$ and $E_{\text{person}}$ for both sets of features. We compute the dot products of the "keys" and "values" embeddings, scale and apply softmax to obtain the attention weights for the person and garment features.

\begin{equation}
\text{Attention Weights}_{\text{person}} = \text{softmax} \left(\frac{E_{\text{garment}} \cdot (E_{\text{person}})^T}{\sqrt{C}}\right)
\end{equation}
\begin{equation}
\text{Attention Weights}_{\text{garment}} = \text{softmax} \left(\frac{E_{\text{person}} \cdot (E_{\text{garment}})^T}{\sqrt{C}}\right)
\end{equation}

\subsubsection{Attention-Guided Feature Fusion}
Next, we utilize the computed attention weights to create attention features, which act as weights for fusing the input features. The attention features $A_{\text{person}}$ and $A_{\text{garment}}$ are computed as follows:
\begin{equation}
A_{\text{person}} = \text{Attention Weights}_{\text{person}} \odot F_{\text{person}}
\end{equation}
\begin{equation}
A_{\text{garment}} = \text{Attention Weights}_{\text{garment}} \odot F_{\text{garment}}
\end{equation}
where $\text{Attention Weights}_{\text{person}}$ and $\text{Attention Weights}_{\text{garment}}$ denote the attention weights for the person and garment features, respectively, and $\odot$ represents element-wise multiplication. The resulting attention features have the same dimensions as the input feature maps.

\subsubsection{Decoder for Final Try-On Result}
The attention-guided features $A_{\text{person}}$ and $A_{\text{garment}}$ are then concatenated and passed through a shallow decoder to generate the final try-on result $T \in \mathbb{R}^{C \times H \times W}$. The shallow decoder is implemented as follows:
\begin{equation}
T = \text{Decoder} \left( [A_{\text{garment}} \otimes A_{\text{person}}]\right)
\end{equation}
where, $[\cdot]$ denotes the concatenation operation, and $\text{Decoder}$ is a shallow neural network responsible for producing the final try-on result.

\subsection{Joint learning Module : Cloth Warping and Tryon}

The warping module is the backbone of the tryon methodology and the results obtained from this stage play a crucial role in the tryon module moving forward.

The key idea behind this constraint is to  integrate explicit learning of the structural and perceptual complexity of the warped garment into our pipeline. The availability of high quality garment segmentation is one of the key factors that motivated us to formulate a learning constraint with the warped garment directly into the single stage pipeline \cite{bai2022single}.
This results into a joint learning of the warped garment and tryon results, improving the results in a parallel manner.

The WCLM uses the parser based label in the form of garment segmentation GT as the guiding parameter. The model is optimised utilising an L1 loss.

The loss $L_{warp}$ is a combination of $I_{wgt}$ and $I_{warp}$ and can be written as:

\begin{equation}
\mathcal{L}_{warp} = ||\bm{I}_{wgt} - \bm{I}_{warp} ||_1,
\end{equation}




\subsection{Training Loss functions}

The L1 loss is a combination of $L_{tryon}$ and $L_{warp}$ mathematically formulated as ,

\begin{equation}
\mathcal{L}_{L1} = ||\bm{I}_{p\_output} - \bm{I}_{p} ||_1 + \mathcal{L}_{warp}
\end{equation}

\par The perceptual loss is based on the VGG network which calculates the L1 distance between the extracted feature maps by the network for both the tryon and warped garment, formulated as ,

\begin{equation}
\mathcal{L}_{prec} = \sum_{i=1}^5 ||\phi_i (\bm{I}_{p\_output}) - \phi_i(\bm{I}_{p}) ||_1,
\end{equation}

\par The style Loss is the optimised statistical error between the feature maps of predicted and GT , where the gram matrix measures the feature correlation between the two inputs,

\begin{equation}
\mathcal{L}_{style} = \sum_i ||G_i^\phi (\bm{I}_{p\_output}) - G_i^\phi\bm{I}_{p}) ||_1,
\end{equation}

where $G_i^\phi$ denotes the Gram matrix of features. 

\par The overall loss is presented as:
\begin{equation}
\mathcal{L} = \sum_{n=1}^N(n+1)*(\lambda_{L1} \mathcal{L}_{L1}^n + \lambda_{prec} \mathcal{L}_{prec}^n + \lambda_{style} \mathcal{L}_{style}^n)
\end{equation}
where $\mathcal{L}^n$ is the loss of the $n_{th}$ scale, and the scale closer to the output is given a larger weight $n-1$.

\section{EXPERIMENTS}

\subsection{Datasets}
The experiments were conducted on two commonly used and publicly available datasets, VITON\cite{Han2017VITONAI} and VITON-HD\cite{Choi2021VITONHDHV}. VITON consists of 14221 training images and a test set of 2032 pairs of images. The view for the images is frontal and paired with the in-shop garment cloth for tryon. VITON-HD contains high-resolution paired images of in-shop garments and their corresponding human models wearing the garments consisting of consists of 11647 train images and 2032 test images.

\begin{table}[t]
    \centering
    \small\addtolength{\tabcolsep}{-1pt}
    \begin{tabular}{c|c|c|c|c}
    \toprule
         Methods & Parser & Warping & FID$\downarrow$ & SSIM $\uparrow$ \\
         \midrule
         VTON \cite{Han2017VITONAI}& Y & TPS & 55.71 & 0.74\\
         CP-VTON \cite{wang2018toward}& Y & TPS & 24.45 & 0.72\\
         CP-VTON++ \cite{minar2020cp}& Y & TPS & 21.04 & 0.75\\
         Cloth-flow\cite{Han2019ClothFlowAF}& Y & AF & 14.43 & 0.84\\ 
         ACGPN\cite{Yang_2020_CVPR}& Y & TPS & 16.64 & 0.84\\
         DCTON\cite{ge2021disentangled}& Y & TPS & 14.82 & 0.83\\
         PF-AFN\cite{ge2021parser} & N & AF & 10.09 & 0.89\\
         $\text{Cloth-flow}^{\star}$\cite{Han2019ClothFlowAF}& N & AF & 10.73 & 0.89\\
         \midrule
         \textbf{Ours}
& N & AF & \textbf{9.14} & \textbf{0.90}\\
         \bottomrule
    \end{tabular}
    \vspace{0.2cm}
    \caption{Quantitative analysis of distinct models evaluated on the VITON dataset. The column "Parser" denotes whether human parsing is employed during model inference, while the column "Warping" indicates the specific warping techniques utilized in the respective models. TPS: Thin Plate Spline. AF: Appearance Flow. $\star$: re-trained with parser free training paradigm.}
    \label{tab:tab1}
\end{table}

\subsection{Implementation details}
All the experiments have been conducted on two A5000 GPUs using PyTorch. The deformable flow is learned the value of $K = 6$. Adam optimizer is used for the batchsize of 8. The model is trained for a learning rate of 0.000035 for 250 epochs. The predefined weights for the loss functions are $\lambda_{L1}=1, \lambda_{prec}=1, \lambda_{style}=100$. The features are extracted for $N=5$ scales with $[64,128,256,256,256]$ number of filters.

\begin{table}[t]
\centering
\centering
\begin{tabularx}{0.48\textwidth}{p{2.3cm}|X<{\centering}X<{\centering}}
\toprule[1.5pt]
Config & SSIM & PSNR \\
\midrule[1pt]
Baseline & 0.817 & 20.78 \\
+LAF & 0.835 & 21.84 \\
+SCFA & 0.863 & 22.43 \\
+Warp Loss & \textbf{0.90} & \textbf{26.5} \\
\bottomrule[1.5pt]
\end{tabularx}
\vspace{0.2cm}
\caption{The effect of different modules.}
\label{tab:tb2}
\end{table}

\begin{table}
    \centering
    \resizebox{\linewidth}{!}{
        \begin{tabular}{@{}lllllll@{}}
            \toprule
            Method & FID$_\text{u}$ $\downarrow$ & KID$_\text{u}$ $\downarrow$ & FID$_\text{p}$ $\downarrow$ & KID$_\text{p}$ $\downarrow$ & SSIM$_\text{p}$ $\uparrow$ & LPIPS$_\text{p}$ $\downarrow$ \\
            \midrule
            VITON-HD~\cite{Choi2021VITONHDHV}        & 14.64 & 6.10 & 12.81 & 5.52 & 0.848 & 0.1216  \\
            HR-VITON~\cite{lee2022hrviton}        & 12.15 & 3.42 & 9.92  & 3.06 & 0.860 & 0.1038  \\
            GP-VTON~\cite{xie2023gpvton}          & 10.49 & 2.23 & 7.71  & 2.01 & 0.857 & 0.0897  \\
            DCI-VTON~\cite{gou2023taming}        & 11.14 & 3.35 & 8.19  & 2.93 & 0.875 & 0.0816  \\
            \midrule 
            \textbf{Proposed}      & \textbf{8.93}  & 1.37 & \textbf{5.60}  & \textbf{0.83} & \textbf{0.877} & \textbf{0.0803}  \\
            \bottomrule
        \end{tabular}
    }
    \caption{Quantitative results on VITON-HD~\cite{Choi2021VITONHDHV}. The subscripts 'u' and 'p' respectively represent the unpaired setting and paired setting.}
    \label{Table:VITONHD}
\end{table}

\subsection{Qualitative Results}

Our method was qualitatively evaluated against CP-VTON+ \cite{minar2020cp}, ACPGN \cite{Yang_2020_CVPR}, PFAPN \cite{ge2021parser}, SDAFN \cite{bai2022single} (Fig. \ref{fig:q1}). While CP-VTON+ and ACPGN are parser-based and prone to misalignment issues, PF-AFN, a parser-free framework, avoids misalignment but suffers from blurred body parts like arms and shoulders.



All methods nearly align the distorted garment with the wearer, but artifacts become more visible as the texture and structural complexity increases. Our model preserves image contrast and skin texture, especially neck region, much better. In Fig. \ref{fig:q1}, our model exhibits superior preservation in left shoulder garment and realistic arm generation, while ensuring structural integrity with minimal misalignments in sleeves and collar edges.

\subsection{Quantitative Results}

The baseline of the proposed approach incorporated a simple warp module and flow without an attention module with the computed scores as shown in table\ref{tab:tb2}, as an improvement we introduce LAF, SCFA and a Warp loss. The LAF module improves the flow which consecutively improves the garment warping, SCFA module encapsulates the attention mechanism’s capability to capture semantic information while considering the context between the garment and the person during the try-on process, and Finally, the warped loss improves the warping by not allowing the learning of warping in bad garment regions as shown in table\ref{tab:tb2}.
Proposed method's assessment on VITON Dataset as shown in Table \ref{tab:tab1}, achieves outstanding results in terms of SSIM and FID with scores of 0.90 and 9.14, respectively, outperforming other methods as shown in Table\ref{tab:tab1}. 
The evaluated scored on VITON-HD dataset also outweigh the performance of proposed method upon comparison with other methods as shown in table\ref{Table:VITONHD} demonstrating the effectiveness of our proposed single-stage framework.

\vspace{-0.1cm}
\section{Conclusion}
We proposed a novel single-stage framework for image-based virtual try-on, achieving photo-realistic fitting results by aligning in-shop garments with clothed person images. Our approach jointly learns warped garments and flow fields using the semantic-contextual fusion attention module and a lightweight linear attention framework. The Warped Cloth Learning Module allows simultaneous learning of warped garment and try-on results. Experiments on the VITON dataset demonstrated state-of-the-art performance both qualitatively and quantitatively. Our method eliminates multi-stage processing, offering a more efficient and reliable virtual try-on experience. The proposed framework shows potential for real-world applications, offering a streamlined virtual try-on process. The future work shall include enhancing robustness on diverse datasets and exploring real-time applications.


\bibliographystyle{IEEEbib}
\small
\bibliography{icme2023template}

\begin{thebibliography}{10}

\bibitem{minar2020cp}
Matiur~Rahman Minar, Thai~Thanh Tuan, Heejune Ahn, Paul Rosin, and Yu-Kun Lai,
\newblock ``Cp-vton+: Clothing shape and texture preserving image-based virtual try-on,''
\newblock in {\em CVPR Workshops}, 2020, vol.~3, pp. 10--14.

\bibitem{Issenhuth2020DoNM}
Thibaut Issenhuth, J{\'e}r{\'e}mie Mary, and Cl{\'e}ment Calauz{\`e}nes,
\newblock ``Do not mask what you do not need to mask: a parser-free virtual try-on,''
\newblock {\em ArXiv}, vol. abs/2007.02721, 2020.

\bibitem{fang2023pg}
Naiyu Fang, Lemiao Qiu, Shuyou Zhang, Zili Wang, and Kerui Hu,
\newblock ``Pg-vton: A novel image-based virtual try-on method via progressive inference paradigm,''
\newblock {\em arXiv preprint arXiv:2304.08956}, 2023.

\bibitem{Choi2021VITONHDHV}
Seunghwan Choi, Sunghyun Park, Min~Gi Lee, and Jaegul Choo,
\newblock ``Viton-hd: High-resolution virtual try-on via misalignment-aware normalization,''
\newblock {\em 2021 IEEE/CVF Conference on Computer Vision and Pattern Recognition (CVPR)}, pp. 14126--14135, 2021.

\bibitem{Han2017VITONAI}
Xintong Han, Zuxuan Wu, Zhe Wu, Ruichi Yu, and Larry~S. Davis,
\newblock ``Viton: An image-based virtual try-on network,''
\newblock {\em 2018 IEEE/CVF Conference on Computer Vision and Pattern Recognition}, pp. 7543--7552, 2017.

\bibitem{wang2018toward}
Bochao Wang, Huabing Zhang, Xiaodan Liang, Yimin Chen, Liang Lin, and Meng Yang,
\newblock ``Toward characteristic-preserving image-based virtual try-on network,''
\newblock {\em ArXiv}, vol. abs/1807.07688, 2018.

\bibitem{Neuberger_2020_CVPR}
Assaf Neuberger, Eran Borenstein, Bar Hilleli, Eduard Oks, and Sharon Alpert,
\newblock ``Image based virtual try-on network from unpaired data,''
\newblock in {\em Proceedings of the IEEE/CVF Conference on Computer Vision and Pattern Recognition (CVPR)}, June 2020.

\bibitem{ge2021disentangled}
Chongjian Ge, Yibing Song, Yuying Ge, Han Yang, Wei Liu, and Ping Luo,
\newblock ``Disentangled cycle consistency for highly-realistic virtual try-on,''
\newblock {\em 2021 IEEE/CVF Conference on Computer Vision and Pattern Recognition (CVPR)}, pp. 16923--16932, 2021.

\bibitem{He2022StyleBasedGA}
Sen He, Yi-Zhe Song, and Tao Xiang,
\newblock ``Style-based global appearance flow for virtual try-on,''
\newblock {\em 2022 IEEE/CVF Conference on Computer Vision and Pattern Recognition (CVPR)}, pp. 3460--3469, 2022.

\bibitem{bai2022single}
Shuai Bai, Huiling Zhou, Zhikang Li, Chang Zhou, and Hongxia Yang,
\newblock ``Single stage virtual try-on via deformable attention flows,''
\newblock in {\em European Conference on Computer Vision}, 2022.

\bibitem{adavala2023generation}
Kiran~Mayee Adavala,
\newblock ``Generation of “hand-drawn” images using deep convolutional generative adversarial networks,''
\newblock {\em resmilitaris}, vol. 13, no. 3, pp. 2104--2111, 2023.

\bibitem{NIPS2015_33ceb07b}
Max Jaderberg, Karen Simonyan, Andrew Zisserman, and koray kavukcuoglu,
\newblock ``Spatial transformer networks,''
\newblock in {\em Advances in Neural Information Processing Systems}, C.~Cortes, N.~Lawrence, D.~Lee, M.~Sugiyama, and R.~Garnett, Eds. 2015, vol.~28, Curran Associates, Inc.

\bibitem{dong2019towards}
Haoye Dong, Xiaodan Liang, Xiaohui Shen, Bochao Wang, Hanjiang Lai, Jia Zhu, Zhiting Hu, and Jian Yin,
\newblock ``Towards multi-pose guided virtual try-on network,''
\newblock in {\em Proceedings of the IEEE/CVF international conference on computer vision}, 2019, pp. 9026--9035.

\bibitem{zamir2022restormer}
Syed~Waqas Zamir, Aditya Arora, Salman Khan, Munawar Hayat, Fahad~Shahbaz Khan, and Ming-Hsuan Yang,
\newblock ``Restormer: Efficient transformer for high-resolution image restoration,''
\newblock in {\em Proceedings of the IEEE/CVF Conference on Computer Vision and Pattern Recognition (CVPR)}, June 2022, pp. 5728--5739.

\bibitem{Han2019ClothFlowAF}
Xintong Han, Weilin Huang, Xiaojun Hu, and Matthew~R. Scott,
\newblock ``Clothflow: A flow-based model for clothed person generation,''
\newblock {\em 2019 IEEE/CVF International Conference on Computer Vision (ICCV)}, pp. 10470--10479, 2019.

\bibitem{Yang_2020_CVPR}
Han Yang, Ruimao Zhang, Xiaobao Guo, Wei Liu, Wangmeng Zuo, and Ping Luo,
\newblock ``Towards photo-realistic virtual try-on by adaptively generating-preserving image content,''
\newblock in {\em IEEE/CVF Conference on Computer Vision and Pattern Recognition (CVPR)}, June 2020.

\bibitem{ge2021parser}
Yuying Ge, Yibing Song, Ruimao Zhang, Chongjian Ge, Wei Liu, and Ping Luo,
\newblock ``Parser-free virtual try-on via distilling appearance flows,''
\newblock {\em arXiv preprint arXiv:2103.04559}, 2021.

\bibitem{lee2022hrviton}
Sangyun Lee, Gyojung Gu, Sunghyun Park, Seunghwan Choi, and Jaegul Choo,
\newblock ``High-resolution virtual try-on with misalignment and occlusion-handled conditions,''
\newblock {\em arXiv preprint arXiv:2206.14180}, 2022.

\bibitem{xie2023gpvton}
Xie Zhenyu, Huang Zaiyu, Dong Xin, Zhao Fuwei, Dong Haoye, Zhang Xijin, Zhu Feida, and Liang Xiaodan,
\newblock ``Gp-vton: Towards general purpose virtual try-on via collaborative local-flow global-parsing learning,''
\newblock in {\em Proceedings of the IEEE/CVF Conference on Computer Vision and Pattern Recognition (CVPR)}, June 2023.

\bibitem{gou2023taming}
Junhong Gou, Siyu Sun, Jianfu Zhang, Jianlou Si, Chen Qian, and Liqing Zhang,
\newblock ``Taming the power of diffusion models for high-quality virtual try-on with appearance flow,''
\newblock in {\em Proceedings of the 31st ACM International Conference on Multimedia}, 2023.

\end{thebibliography}

\end{document}